\documentclass[10pt,twocolumn,letterpaper]{article}

\usepackage{cvpr}
\usepackage[colorlinks=true,citecolor=green]{hyperref}
\usepackage{times}
\usepackage{epsfig}
\usepackage{graphicx}
\usepackage{amsmath}
\usepackage{amssymb}
\usepackage{subfig}
\usepackage{tabularx}
\usepackage[dvipsnames]{xcolor}
\usepackage{algorithm}
\usepackage[noend]{algpseudocode}

\newcommand{\footlabel}[2]{%
    \addtocounter{footnote}{1}%
    \footnotetext[\thefootnote]{%
        \addtocounter{footnote}{-1}%
        \refstepcounter{footnote}\label{#1}%
        #2%
    }%
    $^{\ref{#1}}$%
}

\newcommand{\footref}[1]{%
    $^{\ref{#1}}$%
}

\makeatletter
\renewcommand\@makefnmark{\hbox{\@textsuperscript{\normalfont\color{red}\@thefnmark}}}
\makeatother

\cvprfinalcopy

\begin{document}

%%%%%%%%% TITLE
\title{Images Don't Lie: Transferring Deep Visual Semantic Features to Large-Scale Multimodal Learning to Rank}

\author{Corey Lynch\\
Etsy\\
{\tt\small clynch@etsy.com}
% For a paper whose authors are all at the same institution,
% omit the following lines up until the closing ``}''.
% Additional authors and addresses can be added with ``\and'',
% just like the second author.
% To save space, use either the email address or home page, not both
\and
Kamelia Aryafar\\
Etsy \\
{\tt\small karyafar@etsy.com}
\and
Josh Attenberg\\
Etsy \\
{\tt\small jattenbreg@etsy.com}
}

\maketitle
%\thispagestyle{empty}

%%%%%%%%% ABSTRACT
\begin{abstract}
Search is at the heart of modern e-commerce. As a result, the task of ranking search results automatically (learning to rank) is a multibillion dollar machine learning problem. Traditional models optimize over a few hand-constructed features based on the item's text. In this paper, we introduce a multimodal learning to rank model that combines these traditional features with visual semantic features transferred from a deep convolutional neural network. In a large scale experiment using data from the online marketplace Etsy~\footlabel{etsyurl}{\url{www.etsy.com}}, we verify that moving to a multimodal representation significantly improves ranking quality. We show how image features can capture fine-grained style information not available in a text-only representation. In addition, we show concrete examples of how image information can successfully disentangle pairs of highly different items that are ranked similarly by a text-only model.
\end{abstract}

%%%%%%%%% BODY TEXT
\section{Introduction}
\label{sec:intro}
%%%intro

\begin{figure}[htpb]
\centering
\includegraphics[scale=1]{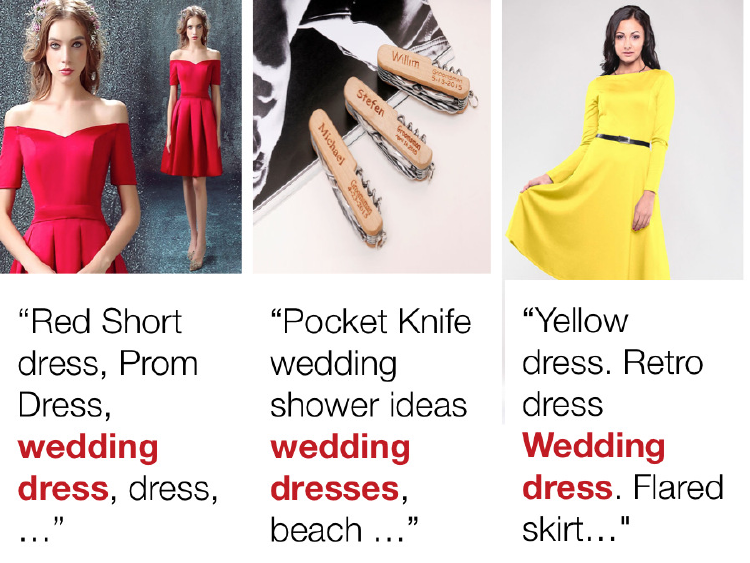}
\caption{\textbf{Irrelevant search results for the query ``wedding dress":} Even though it's apparent in the images that these are not wedding dresses, each listing's descriptive title contains the phrase ``wedding dress", allowing it to show in search results for the query.}
\label{fig:motivation}
\end{figure}

Etsy~\footref{etsyurl} is a global marketplace where people buy and sell unique goods: handmade items, vintage goods, and craft supplies. Users come to Etsy to search for and buy listings other users offer for sale. A \textit{listing} on Etsy consists of an image of the item for sale along with some text describing it. With 35 million listings for sale~\footlabel{etsystats}{The statistics reported in this paper are accurate at the time of submission of this paper.}, correctly ranking search results for a user's query is Etsy's most important problem. Currently Etsy treats the ranking problem as an example of supervised learning: learn a query-listing relevance function from data with listing feature representations derived from listings' titles and tags. However, with over 90 million listing images, we wonder: \textit{is there useful untapped information hiding in Etsy's images that isn't well captured by the descriptive text?} If so, how can we integrate this new data modality into Etsy's existing ranking models in a principled way? In this paper we attempt to explore these two major questions on a real world dataset containing over $1.4$ million Etsy listings with images. Specifically, we describe a multimodal learning to rank method which integrates both visual and text information. We show experimentally that this new multimodal model significantly improves the quality of search results.

\begin{figure*}[htpb]
\centering
\includegraphics[scale=0.6]{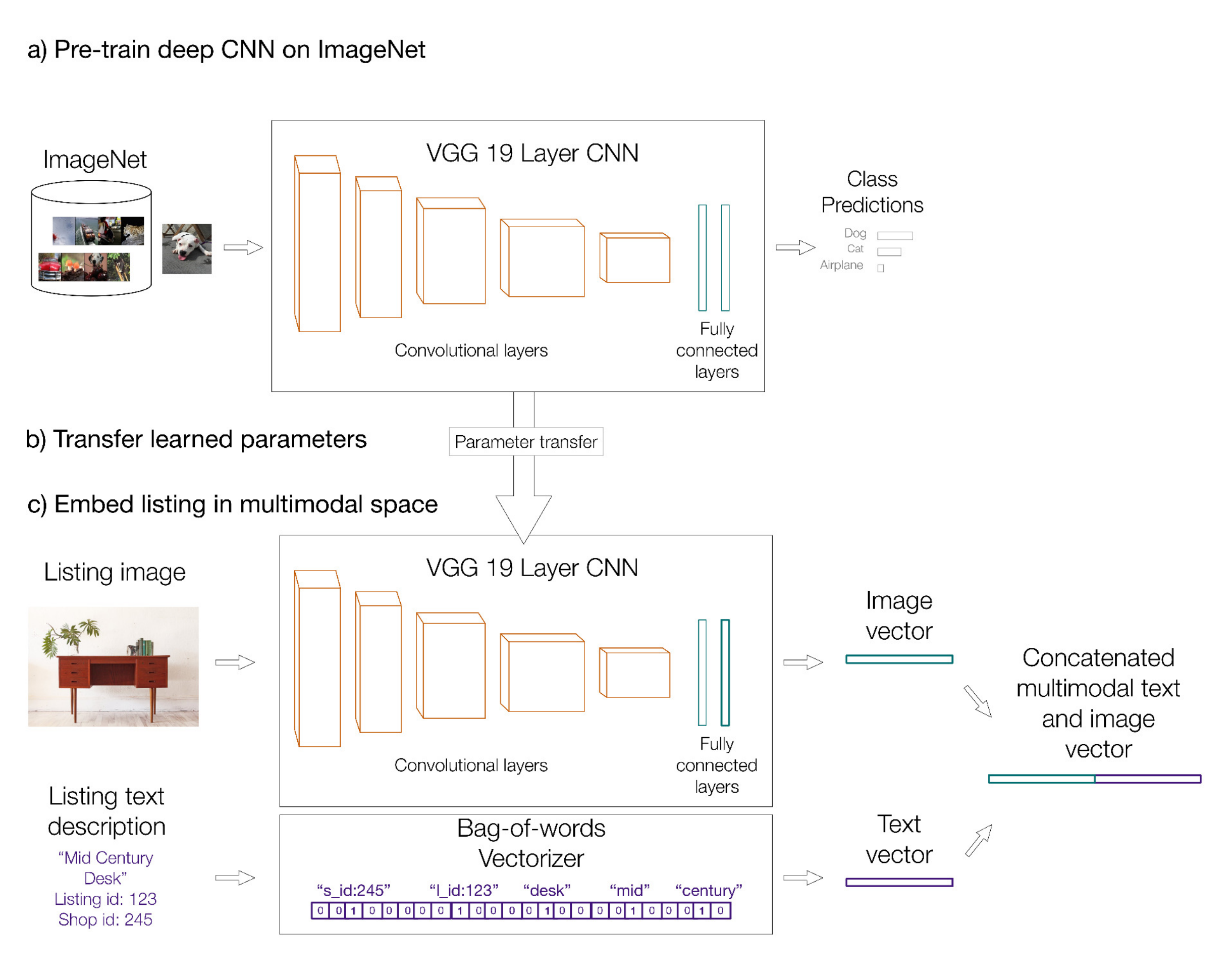}
\caption{\textbf{Transferring parameters of a CNN to the task of multimodal embedding:} In a) we utilize a pre-trained $19$ layer VGG-style network that is trained on a large scale object recognition task (ImageNet challenge). In b) we remove the last layer (containing scores for the different object classes) and transfer the parameters of the modified network to our task. In c) we use the modified network as a fixed feature extractor of high-level image content information, taking the last fully connected layer as an image embedding. We simultaneously embed the listing's text in a bag of words space, then concatenate the two embeddings to form a single multimodal descriptor of a listing.}
\label{fig:mmembedder}
\end{figure*}

We motivate this paper with a real example of a problem in search ranking. Figure~\ref{fig:motivation} shows listings that appear in the results for the query ``wedding dress". Even though it is clear from looking at the images that they aren't wedding dresses, each listing's title contains the term ``wedding dress", causing it to appear in the results. This kind of term noise in listing descriptions is pervasive. Sellers write their own descriptions and are motivated to include high traffic query terms to boost the visibility of their listings in search. It is obvious that there is \textit{complementary information} in the listing images that is either unavailable in the text or is even in contrast to the text. Our hypothesis is that if we can mine this complementary high-level visual information, we can incorporate it into our models to improve search ranking.

With the goal of learning high-level content directly from the image, deep convolutional neural networks (CNN) are an obvious model choice. 
CNNs are a powerful class of models inspired by the human visual cortex that rivals human-level performance on difficult perceptual inference tasks such as object recognition~\cite{krizhevsky2012imagenet}. \cite{zeiler2014visualizing} shows that image feature representations learned on large scale object recognition tasks are powerful and highly interpretable. The lower layers of the network learn low-level features like color blobs, lines, corners; middle layers combine lower layers into textures; higher layers combine middle layers into higher-level image content like objects. High-level visual information is made increasingly explicit along the model's processing hierarchy~\cite{gatys2015texture}. This high-level description formed in the deep layers of the network is what we are interested in mining as a rival source of information to the listing’s text description.

One consideration to make is that large modern CNNs require large amounts of training data~\cite{krizhevsky2012imagenet,srivastava2014dropout}. Even though Etsy's production search system generates millions of training examples per day in aggregate, the amount of examples available to an individual query model can be in the low hundreds, particularly for queries in the long tail. This makes training one deep CNN per query from scratch prone to overfitting. Transfer learning is a popular method for dealing with this problem, with many examples in computer vision~\cite{pan2010survey,aytar2011tabula,tommasi2014learning}. 
We take the pre-trained 19-layer VGG net ~\cite{simonyan2014very} as a fixed extractor of general high-level image features. We chose this model due to it’s impressive performance on a difficult 1000-way object classification task. Our assumption is that the activations of it’s neurons immediately prior to the classification task contain general high-level information that may be useful to our ranking task. Figure~\ref{fig:mmembedder} gives a high-level overview of our multimodal feature extraction process.

Learning to rank search results has received considerable attention over the past decade~\cite{burges2005learning,bai2010learning,joachims2002optimizing,hang2011short}, and it is at the core of modern information retrieval. A typical setting of learning to rank for search is to: (i) embed documents in some feature space, (ii) learn a ranking function for each query that operates in this feature space over documents. Early approaches optimized over a few hand-constructed features, e.g. item title, URL, PageRank ~\cite{joachims2002optimizing}. More recent approaches optimize over much larger sets of orthogonal features based on query and item text ~\cite{bai2010learning}. Our research follows this orthogonal approach and explores the value of a large set of image features. Our baseline listing representation consists of features for a listing's terms, a listing's id, and a listing's shop id. The presence of the latter two features captures historical popularity information at the listing and shop level respectively for the query.

To our knowledge, this is the first investigation of the value of transferring deep visual semantic features to the problem of learning to rank for search. The results of a large-scale learning to rank experiment on Etsy data confirms that moving from a text-only representation to a multimodal representation significantly improves search ranking. We visualize how the multimodal representation provides complementary style information to the ranking models. We also show concrete examples of how pairs of highly different listings ranked similarly by a text model get disentangled with the addition of image information. We feel this, along with significant quantitative improvements in offline ranking metrics demonstrates the value of the image modality.

This paper is organized as follows: In Section~\ref{sec:method} we describe our multimodal ranking framework. Section~\ref{subsec:mmembedding} gives a detailed explanation of how we obtain multimodal embeddings for each listing. Section~\ref{subsec:ltr} gives a brief introduction to learning to rank and describes how these embeddings are incorporated into a pairwise learning to rank model. Section~\ref{sec:results} describes a large scale experiment where we compare multimodal models to text models. Finally in Section~\ref{sec:conclusion}, we discuss how moving to a multimodal representation affects ranking with qualitative examples and visualizations.
%-------------------------------------------------------------------------

\section{Methodology}
\label{sec:method}
Here we describe how we extend our existing learning to rank models with image information. We first explain how we embed listings for the learning to rank task in both single modality and multimodal settings. Then we explain how listings embedded in both modalities are incorporated into a learning to rank framework.

\subsection{Multimodal Listing Embedding}
\label{subsec:mmembedding}
%%%mm subsection
Each Etsy listing contains text information such as a descriptive title and tags, as well as an image of the item for sale. To measure the value of including image information, we embed listings in a multimodal space (consisting of high-level text and image information), then compare the multimodal representation to the baseline single modality model. Let $\mathbf{d}_i$ denote each listing document. We then use $\mathbf{d}_{T_i} \in \mathbb{R}^{|T|}$ and $\mathbf{d}_{I_i} \in \mathbb{R}^{|I|}$ to denote the text and image representation of $\mathbf{d}_i$ respectively. $|T|$ denotes the dimensionality of the traditional feature space which is a sparse representation of term occurrences in each $\mathbf{d}_i$. $|I|$ denotes the dimensionality of the image feature space which in contrast to text is a dense feature space. The goal of multimodal embedding is then to represent a listing through text and image modalities in a single vector, i.e. the $\mathbf{d}_ {MM_i} \in \mathbb{R}^{|M|}$ where $|M|$ is the dimensionality of the final embedding.

\def\0{{\bf 0}}

\begin{algorithm}
\caption{Multimodal Embedding of Listings}
\begin{algorithmic}[1]
\Procedure{EmbedMultimodal}{$\mathbf{d}_i$ }
\State $ \mathbf{d}_{T_i} \gets \textbf{BoW}(text)$
\State $ \mathbf{d}_{I_i} \gets \textbf{VGG}(image)$ 
\State $ \mathbf{d}_{MM_i} \gets [\mathbf{d}_{T_i} ,\mathbf{d}_{T_i} ]$
\State \textbf{return} $\mathbf{d}_{MM_i}$
\EndProcedure
\end{algorithmic}
\label{alg:mmembed}
\end{algorithm}

\newcommand{\tuple}[1]{\ensuremath{\left \langle #1 \right \rangle }}

\begin{algorithm}
\caption{Generate Pairwise Classification Instances}
\begin{algorithmic}[1]
\Procedure{GetPairwiseInstances}{$\{\tuple{\mathbf{d}_i^{+},\mathbf{d}_i^{-}}\}$ }
\State $L \gets \{\}$
\For{$i=1\dots |P|$}\Comment{$|P|$ labeled tuples}
\State $\mathbf{d}^{+} _ {MM_i} \gets \textbf{EmbedMultimodal}(\mathbf{d}_i^{+})$ 
\State $\mathbf{d}^{-} _ {MM_i} \gets \textbf{EmbedMultimodal}(\mathbf{d}_i^{-})$ 
\State Draw $r$ uniformly at random from $[0,1)$
\If {$r > 0.5$}
\State $x_i \gets \mathbf{d}^{+}_{MM_i} - \mathbf{d}^{-}_{MM_i}$
\State $y_i \gets +1$
\Else  
\State $x_i \gets \mathbf{d}^{-}_{MM_i} - \mathbf{d}^{+}_{MM_i}$
\State $y_i \gets -1$
\EndIf
\EndFor
\State $L = L.append(\tuple{x_i,y_i})$
\State \textbf{return} $L$\Comment{The list of classification instances.}
\EndProcedure
\end{algorithmic}
\label{alg:generateInstance}
\end{algorithm}

\begin{figure}[htpb]
\centering
\includegraphics[scale= 0.3,trim=4 4 4 4,clip]{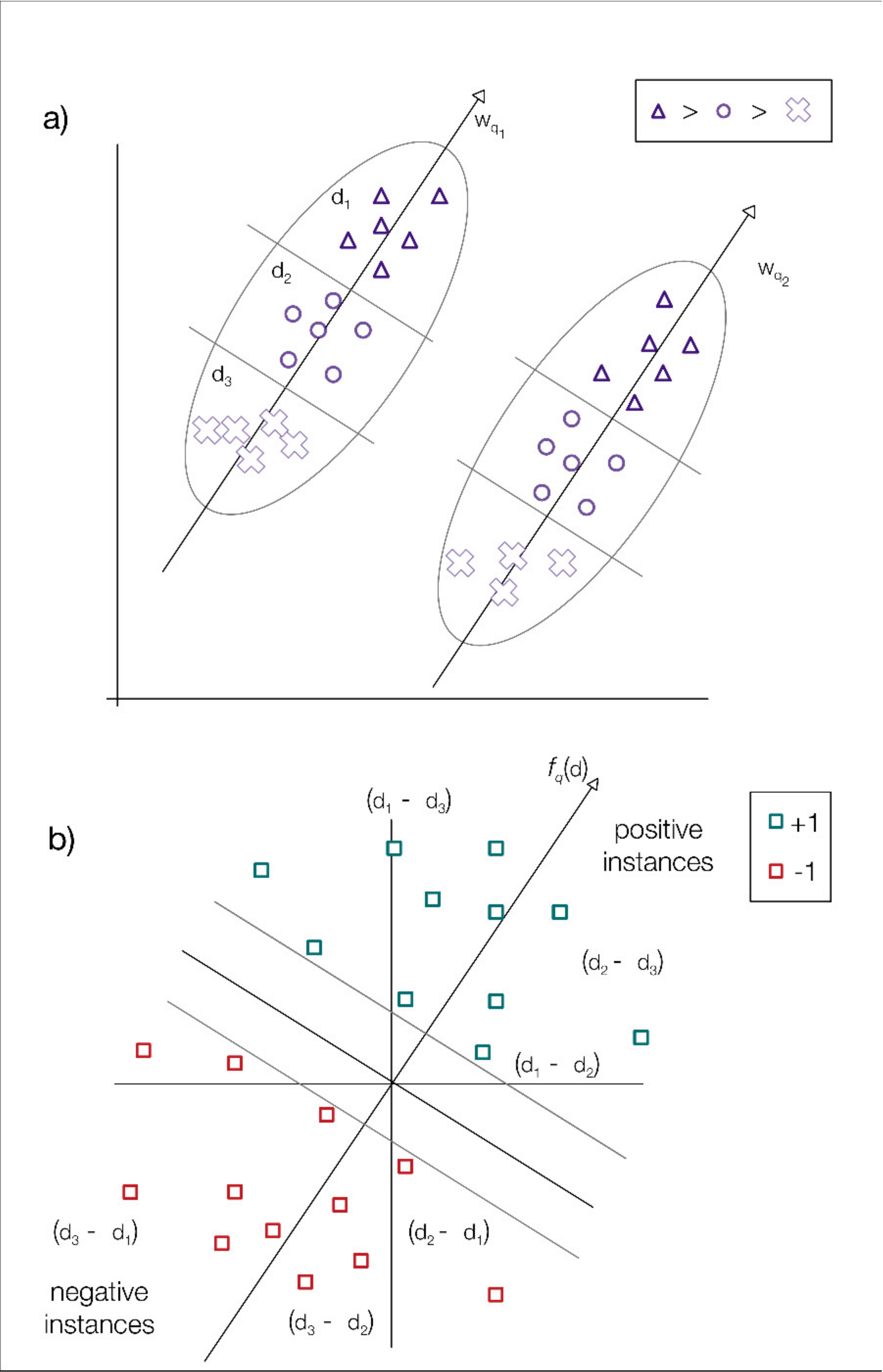}
\caption{\textbf{Transformation to Pairwise Classification~\cite{hang2011short}:} a) Shows a synthetic example of the ranking problem. There are two groups of documents (associated with different queries) embedded in some feature space. Documents within each group have different relevance grades: $relevance(\mathbf{d}_1) > relevance(\mathbf{d}_2) > relevance(\mathbf{d}_3)$. The weight vector $w$ corresponds to a linear ranking function $f(\mathbf{d}) = \tuple{w,\mathbf{d}}$ which can score and rank documents. Ranking documents with this function is equivalent to projecting the documents onto the vector and sorting documents according to the projections. A good ranking function is one where documents in $\mathbf{d}_1$ are ranked higher than documents in $\mathbf{d}_3$, and so on. Documents belonging to different query groups are incomparable. b) shows how the ranking problem in (a) can be transformed into a pairwise classification task: separate well-ordered and non-well ordered pairs. In this task, well-ordered pairs are represented as the vector difference between more relevant and less relevant document vectors,  e.g., $\mathbf{d}_1 - \mathbf{d}_2$, $\mathbf{d}_1 - \mathbf{d}_3$, and $\mathbf{d}_2 - \mathbf{d}_3$. Non-well ordered-pairs are represented as the vector difference between less relevant and more relevant document vectors,  e.g., $\mathbf{d}_3 - \mathbf{d}_1$, $\mathbf{d}_2 -\mathbf{d}_1$, and $\mathbf{d}_3 - \mathbf{d}_2$. We label well-ordered instances $+1$, non-well-ordered $-1$, and train a linear SVM, $f_q(\mathbf{d})$, which separates the new feature vectors. The weight vector $w$ of the SVM classifier corresponds to the ranking function $w$ in (a).}
\label{fig:transformRankingClassification}
\end{figure}

Given a listing document $\mathbf{d}_i$ consisting of the title and tag words, a numerical listing id, a numerical shop id, and a listing image, we obtain a multimodal feature vector $\mathbf{d}_{MM_i}$ by embedding traditional terms in a text vector space, embedding the image in an image vector space, then concatenating the two vectors. Algorithm~\ref{alg:mmembed} describes this multimodal embedding. The text-based features are based on the set of title and tag unigram and bigrams, the listing id, and the shop id for each $\mathbf{d}_i$. These baseline features are then represented in a bag-of-words (BoW)~\cite{weinberger2009feature} space as $\mathbf{d}_{T_i} \in \mathbb{R}^{|T|}$. $|T| = |D| + |L| + |S|$, where $|D|$ is the size of the dictionary, $|L|$ is the cardinality of the set of listings, and $|S|$ is the cardinality of the set of shops. Each element in $\mathbf{d}_{T_i}$ is $1$ if $\mathbf{d}_i$ contains the term, and is $0$ otherwise.

Each $\mathbf{d}_i$ is also represented in the image space as $\mathbf{d}_{I_i} \in \mathbb{R}^{|I|}$. To obtain $\mathbf{d}_{I_i}$, we adopt a transfer learning approach. Oquab~\etal~\cite{oquab2014learning} shows that the internal layers of a convolutional neural network pre-trained on ImageNet can act as a generic extractor of a mid-level image representation and then re-used on other tasks. In our model, we utilize the VGG-19 network~\cite{simonyan2014very} and remove the last fully-connected layer. For every listing $\mathbf{d}_i$, we scale it’s image uniformly so that its shortest spatial dimension is $256$ pixels, then take the center $224 \times 224$ crop. We then pass the cropped image through the modified VGG network to get a $4096$ dimensional feature vector~\footnote{We make our scalable VGG-19 feature extractor available at:\\ \url{https://github.com/coreylynch/vgg-19-feature-extractor}}. This vector contains the activations fed to the original object classifier. We $\ell_2$ normalize the activations, then use the normalized vector as the final $4096$ dimensional image representation $\mathbf{d}_{I_i}$. The multimodal representation of the document $\mathbf{d}_i$ can now be obtained simply as $\mathbf{d}_ {MM_i} \in \mathbb{R}^{|M|} = [\mathbf{d}_{T_i}, \mathbf{d}_{I_i}]$ where $|M| = |T| + |I|$ is the dimensionality of the final multimodal representation. Figure~\ref{fig:mmembedder} illustrates this process.

\subsection{Learning To Rank}
\label{subsec:ltr}
%%%%%ltr
E-commerce search is a task that looks like the following: (i) Query: a user comes to the site and enters a query, $q$, e.g. ``desk".
(ii) Retrieval: the search engine finds all listing documents that contain terms from the query in their title or tag descriptions, such as ``mid century desk", ``red desk lamp", etc. (iii) Ranking: the search engine uses a ranking function to score each listing document, where a higher score expresses that a listing document is more relevant to the query. The search engine sorts listings by that score, and presents the results to the user. The goal of learning to rank~\cite{burges2005learning} is to automatically learn (iii), the ranking function, from historical search log data.

\begin{figure*}[htpb]
\centering
\begin{center}
\includegraphics[scale = 0.8]{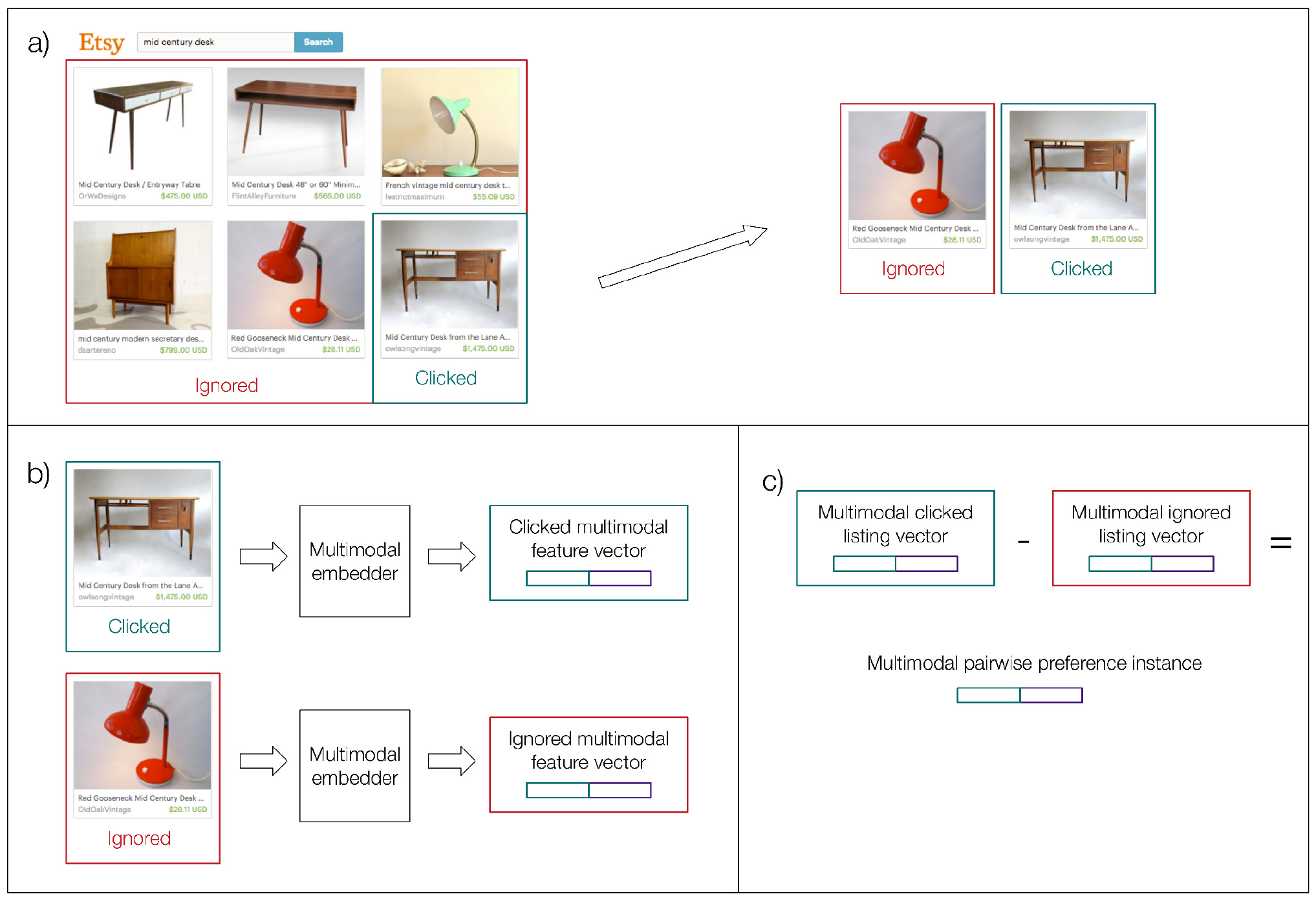}
\end{center}
\caption{\textbf{From search logs to multimodal pairwise classification instances:} a) A user comes to the site, enters the search query \textbf{mid century desk}, is presented with some results, clicks one and ignores the others. We take the listing she clicked and an adjacent listing she ignored as an instance of implicit pairwise preference in the context of the query \textbf{mid century desk} forming a training triplet $(q, \mathbf{d}^+, \mathbf{d}^-)$ from (\textbf{mid century desk}, clicked listing, ignored listing). b) We embed both listings in the pair in multimodal space. c) Map labeled embedded pair to a single pairwise classification instance. We flip a coin. If heads, create a well-ordered pairwise instance (clicked vector - ignored vector) and label it $+1$; if tails, create a non-well-ordered pairwise instance (ignored vector - clicked vector) and label it $-1$.}
\label{fig:pairInstance}
\end{figure*}

In this paper, we restrict our attention to the \textit{pairwise preference} approach to learning a ranking function~\cite{herbrich1999large}. That is, given a set of labeled tuples $P$, where each tuple contains a query $q$, a relevant document $\mathbf{d}^{+}$ and an irrelevant document $\mathbf{d}^{-}$, we want to learn a function $f_q(\mathbf{d})$ such that $f_q(\mathbf{d}^{+}) > f_q(\mathbf{d}^{-})$. In our case, relevance is binary and determined by the user: a user is said to judge a search result relevant if she purchases the listing, adds it to her cart, or clicks on the listing and dwells on it for longer than 30 seconds.

As shown by Herbrich~\etal~\cite{herbrich1999large}, the ranking problem can be transformed into a two-class classification: learn a linear classifier that separates well-ordered pairs from non-well-ordered pairs. This is illustrated in Figure~\ref{fig:transformRankingClassification}. To achieve this, we can transform any implicit relevance judgement pair $(\mathbf{d}^+, \mathbf{d}^-)$ into either a well-ordered or non-well ordered instance. Specifically, suppose for each preference pair $(q, \mathbf{d}^+, \mathbf{d}^-)$ we flip a coin. If heads the preference pair $(q, \mathbf{d}^+, \mathbf{d}^-) \mapsto (\mathbf{d}^+ - \mathbf{d}^-, +1)$ (a well-ordered pair), else $(q, \mathbf{d}^+, \mathbf{d}^-) \mapsto (\mathbf{d}^- - \mathbf{d}^+, -1)$ (a poorly ordered pair). 

This results in an evenly balanced pairwise classification dataset for each query $q$. Algorithm~\ref{alg:generateInstance} explains the process of generating classification instances for input pairwise preference tuples. The new classification task can now be solved by minimizing the regularized hinge classification loss:

\begin{equation}
\min_w \sum_{i=1}^{m} max([1 - y_i \left \langle w,x_i \right \rangle],0) + \lambda_1\left \| w\right \|^1 + \lambda_2\left \| w \right \|^2  \nonumber
\end{equation}

via stochastic gradient descent. A well trained pairwise classifier minimizes the number of pairs which are ranked out of order, i.e. the ranking loss. To rank a new set of listing documents for a query, we embed them in the feature space, then use output of the trained classifier to obtain ranking scores for each. This method, also known as RankingSVM, is used extensively in the ranking literature~\cite{herbrich1999large, burges2005learning, hang2011short}.

Following ~\cite{radlinski2006minimally}, we can obtain large quantities of \textit{implicit} pairwise preference instances cheaply by mining Etsy's search logs\footnote{We obtain labeled pairs for each query using the FairPairs method. A well known and significant problem in collecting training data from search is presentation bias~\cite{radlinski2006minimally}: users click on higher presented results irrespective of query relevance. Ignoring this bias and training on naively collected data can lead to models that just learn the existing global ranking function. The FairPairs method modifies search results in a non-invasive manner that allows us to collect pairs of (preferred listing, ignored listing) that are unaffected by this presentation bias.}. 

\begin{figure}[htpb]
\centering
\includegraphics[scale= 1]{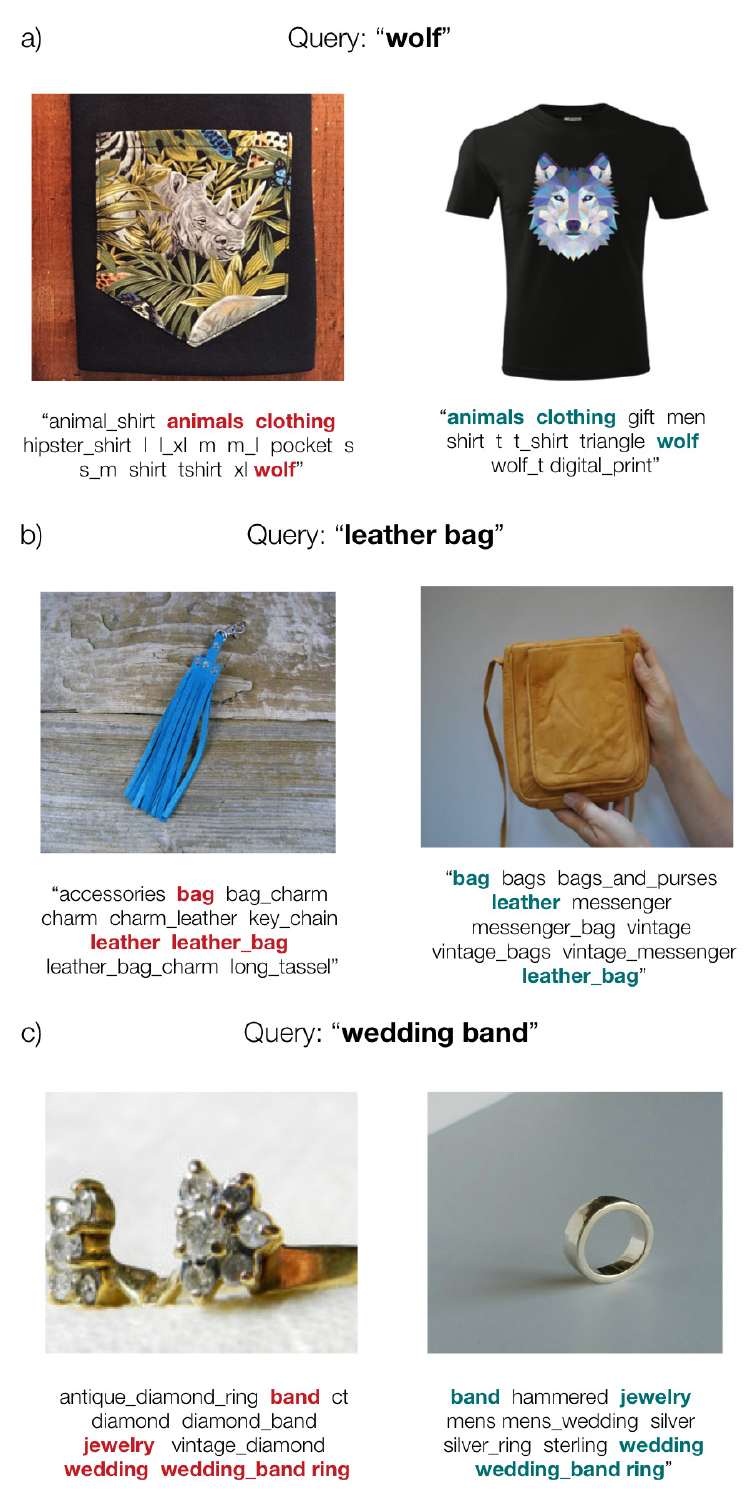}
\caption{\textbf{Image information can help disentangle different listings considered similar by a text model:} Here are three pairs of highly different listings that were ranked similarly by a text model, and far apart by a multimodal model. Title terms that both listings have in common are bolded, showing how text-only models can be confused. In every case, the multimodal model ranks the right listing higher. All three queries benefit from substantial gains in ranking quality (NDCG) by moving to a multimodal representation. In a) for example, we see two listings returned for the query ``wolf'': the left listing's image shows a picture of a rhino, while the right listing's image shows a t-shirt with a wolf on it. A text-only model ranked these listings only 2 positions apart, likely due to the overlapping words in their titles. A multimodal model, on the other hand, ranked the right listing 394 positions higher than the left. The query ``wolf'' saw a $3.07\%$ increase in NDCG by moving to a multimodal representation.}
\label{fig:pathological}
\end{figure}

The process for doing so is as follows: A user comes to the site, enters a query, and is presented with a page of results. If she interacts with listing $\mathbf{d}_i$ and ignores the adjacent listing $\mathbf{d}_j$, a reasonable assumption is that she prefers $\mathbf{d}_i$ over $\mathbf{d}_j$ in the context of query $q$. We call this an implicit relevance judgement, mapping $(q, \mathbf{d}_i, \mathbf{d}_j) \mapsto (q, \mathbf{d}^+, \mathbf{d}^-)$, forming the necessary input for Algorithm~\ref{alg:generateInstance}. Figure~\ref{fig:pairInstance} illustrates how we move from search logs to multimodal pairwise classification instances.

%-------------------------------------------------------------------------

\section{Results and Discussion}
\label{sec:results}
%%%results

This section describes a large scale experiment to determine how a multimodal listing representation impacts ranking quality. In Section~\ref{subsec:metrics}, we describe our ranking quality evaluation metric. In Section~\ref{subsec:dataset} we describe our dataset. Finally we present our findings in Section~\ref{subsec:results}.

\subsection{Evaluation metrics}
\label{subsec:metrics}

To evaluate the quality of one modality ranking model over another, we measure the model's average \textit{Normalized Discounted Cumulative Gain} (NDCG)~\cite{jarvelin2002cumulated} on holdout search sessions. NDCG is the standard measure of a model's ranking quality in information retrieval. It lies in the $[0, 1]$ range, where a higher NDCG denotes a better holdout ranking quality. Here we give a brief background on NDCG and why it is a good choice for quantifying ranking quality.

The cumulative gain (CG) of a ranking model's ordering is the sum of relevance scores over the ranked listings. The CG at a particular rank position p is defined as:

 $$\mathrm{CG_{p}} = \sum_{i=1}^{p} rel_{i},$$ 
where $rel_{i}$ is the \textit{implicit} relevance of the result at position $i$.

CG on its own isn't a particularly good measure of the quality of a model's ordering: moving a relevant document above an irrelevant document does not change the summed value.
\begin{figure*}[htpb]
\centering
\includegraphics[scale = 1.1]{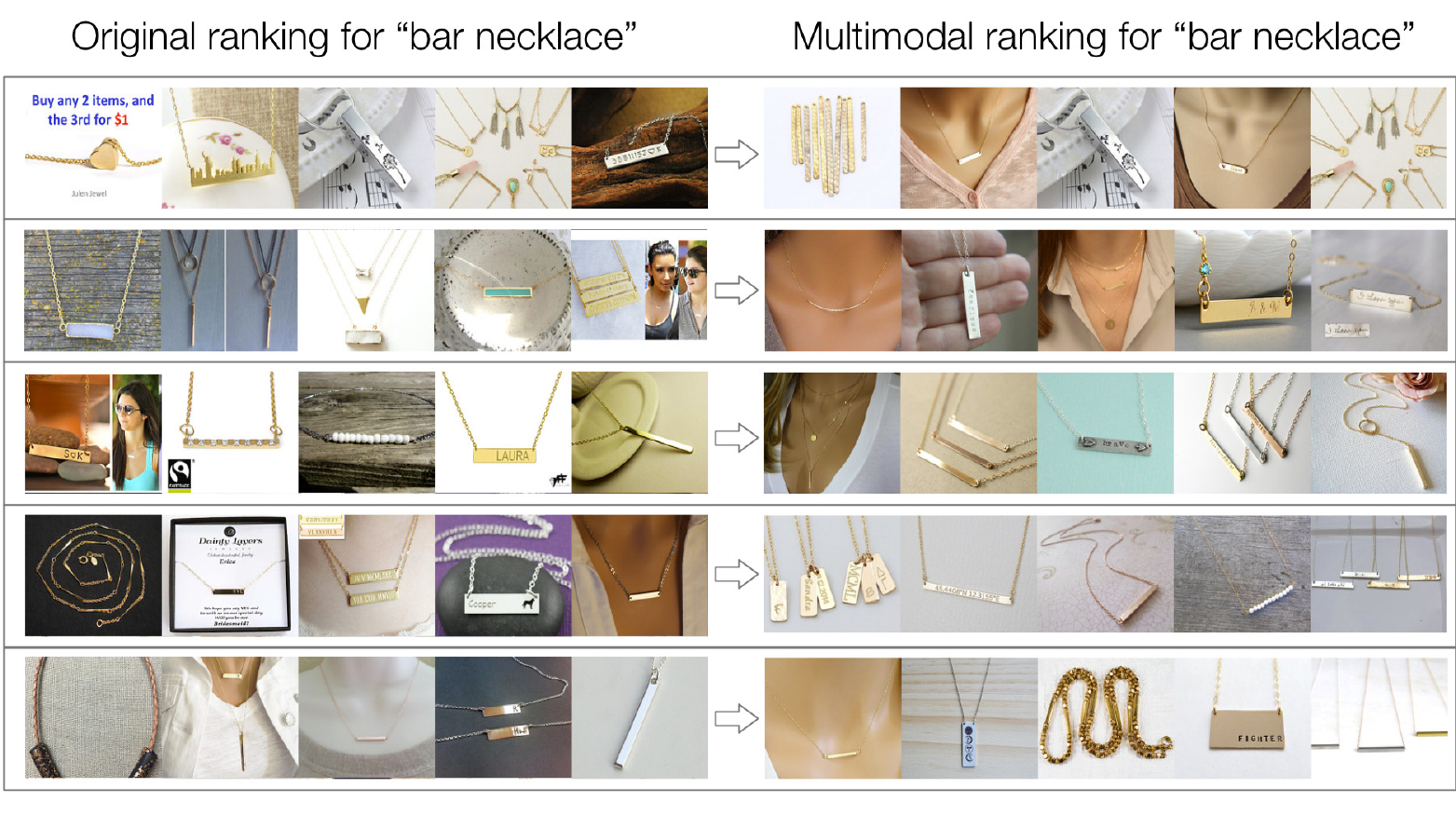}
\caption{\textbf{Visualizing ranking changes by incorporating image information:} Here we visualize how moving from a text-only representation to a multimodal one impacted the ranking for the query ``bar necklace''. Each row shows the top ranked listings for the query by percentile, from $90^{th}$ percentile on the top to $50^{th}$ percentile on the bottom. We observe that the images in each row show more visual consistency under a multimodal ranking than a text-only ranking. That, along with the fact that the query saw a $6.62\%$ increase in ranking quality (NDCG) by moving to a multimodal representation, suggests that the multimodal representation captures relevant visual style information not available in the text.}
\label{fig:continum}
\end{figure*}

This motivates discounted cumulative gain (DCG)~\cite{burges2005learning} as a ranking measure. DCG is the sum of each listing's relevance score \textit{discounted by the position it was shown}. DCG is therefore higher when more relevant listings are ranked higher in results, and lower when they are ranked lower. The DCG is defined as:

$$ \mathrm{DCG_{p}} = \sum_{i=1}^{p} \frac{ 2^{rel_{i}} - 1 }{ \log_{2}(i+1)}. $$  

Finally, we arrive at Normalized Discounted Cumulative Gain (NDCG) by dividing a model's DCG by the ideal DCG (IDCG) for the session:

$$ \mathrm{nDCG_{p}} = \frac{DCG_{p}}{IDCG_{p}} $$

This gives us a number between $0$ and $1$ for the model's ranking of a holdout search result set, where a higher NDCG approaches the ideal DCG, denoting a better ranking quality.

Our holdout data (both validation and test) are in the form of a list of labeled sessions for each query $q$. A labeled session is a page of search results presented to a user for the query $q$, where the user has provided implicit relevance judgements.\footnote{We collect implicit relevance judgments in the test period the same as in training: documents are labeled $0.0$ if ignored and $1.0$ if the user purchased the listing, added the listing to their cart, or clicked on the listing and dwelled for more than $30$ seconds.} We evaluate a trained query model $f_q$ by computing its average NDCG over all labeled holdout sessions for the query $q$. To achieve this, we first compute each session's NDCG as follows: i) embed all the session's listing documents in the model space, ii) score and rank embedded listings with the model, and iii) compute the labeled session NDCG. Then we average over session NDCG to get average query NDCG. Finally, to compare the ranking ability of models based in one modality vs. another, we report the average modality NDCG across query NDCGs.

\subsection{Dataset}
\label{subsec:dataset}

For our experiment, we select a random 2 week period in our search logs to obtain training, validation, and test data. We mine query preference pairs from the first week as \textit{training data}. We mine labeled holdout sessions from the following week, splitting it evenly into \textit{validation} and \textit{test} sessions. This results in $8.82$ million training preference pairs, $1.9$ million validation sessions, and $1.9$ million test sessions. Across training and test there are roughly $1.4$ million unique listing images.

We tune model learning rates, $\lambda_1$ and $\lambda_2$ regularization strengths on our validation data. We also allow queries to select the best modality based on validation NDCG from the set of multiple modalities : \textit{text-only}, \textit{image-only}, and \textit{multimodal} prior to test. Of the $1394$ total queries we built models for, $51.4\%$ saw an increase in validation NDCG by utilizing the multimodal representation over the baseline representation. We present results in Section~\ref{subsec:results}.

\subsection{Results}
\label{subsec:results}
%-------------------------------------------------------------------------
Here we present and discuss the results of incorporating image information into our ranking models. Table~\ref{tab:results} summarizes the results of our experiment for queries that moved from a text-only representation to a multimodal one. For each modality, we present the average lift in NDCG over our baseline text-only models. We find that consistent with our expectations, switching from a text-only ranking modality to one that uses image information as well (MM) yields a statistically significant $\mathbf{1.7\%}$ \textbf{gain in average ranking quality}. We find it interesting that a purely image-based representation, while strictly worse than both other modalities, only underperforms the baseline representation by $2.2\%$.

\begin{table}[htpb]
   \label{tab:results}
   \caption{Lift in Average NDCG, relative to baseline ($\%$), on sample dataset is compared across various modalities. $^{*}$ indicates the change over baseline method is statistically significant according to Wilcoxon signed rank test at the significance level of $0.0001$.}
    \begin{center}
\begin{tabular}{|c||c|c|c|} 
\hline
Modality&Text&Image&MM\\
\hline
 \hline
Relative lift in NDCG& $+0.0\%$ &$-2.2\%^{*}$&$+\mathbf{1.7}\%^{*}$\\ \hline
 \end{tabular}
 \label{tab:results}
   \end{center}
\end{table}

% \begin{figure*}[h!]
% \centering
% \includegraphics[scale = 1.1]{./continuumvis2HigherRez.png}
% \caption{\textbf{Visualizing ranking changes for the query \textbf{bar necklace} by incorporating image information:} Here we visualize how models based on image and text information rank listings differently than models based on text alone. Rows are ranking percentiles, from $50^{th}$ on the bottom to $90^{th}$ on the top. We observe that in each percentile band, the top ranked items under the multimodal model show more visual consistency than the ones ranked by a text-only model. This, along with a $6.62\%$ increase in ranking quality (NDCG) by moving to a multimodal representation, suggests that the multimodal representation captures \textit{relevant visual style information} not available in the text.}
% \label{fig:continum2}
% \end{figure*}

We can explore how incorporating image information changes the ranking by looking at a continuum visualization of how the two modality models rank listings for a query: The top row shows the top 90th percentile of ranked listings, the bottom row shows the 50th percentile of ranked listings, and every row in between is a continuum. Figure~\ref{fig:continum} a continuum visualization for the query bar  ``bar necklace''. We observe that by moving from a text-only to a multimodal representation, the top ranked listings in each band show greater visual consistency. This suggests that there is \textit{relevant style information} captured in the image representation that is not available, or not well described in the text representation. Incorporating this complementary side information leads to a $\mathbf{6.62\%}$ \textbf{increase} in offline NDCG.

We can also see concrete examples of the image representation providing valuable complementary information to the ranking function. Figure~\ref{fig:pathological} shows three examples of listings that were ranked similarly under a text model and very differently under a multimodal model. For example, Figure~\ref{fig:pathological} (a) shows two listings that match the query ``wolf". It is apparent by looking at the two listing images that only the right listing is relevant to the query: the left shows a picture of a rhino; the right is a t-shirt with a wolf on it. The text-only model ranked these two listings only two positions apart. In contrast, the multimodal model ranked the relevant right-hand listing 394 positions higher. We can see by looking at the bolded terms in common how the text model could be confused: as points embedded in text space, these listings are essentially on top of each other. It is clear that the images contain enough information to differentiate them: one contains a wolf, one contains a rhino. By embedding listings in a space that preserves this difference, the multimodal model can effectively disentangle these two listings and provide a more accurate ranking. The query ``wolf" saw an overall $\mathbf{3.07\%}$ \textbf{increase} in NDCG by moving to a multimodal representation.
Similarly for Figure~\ref{fig:pathological} (b), a text model for the query ``leather bag" ranked the two listings 1 position apart, while the multimodal model ranked the right listing 660 positions higher. ``leather bag" saw a $\mathbf{2.56\%}$ \textbf{increase} in NDCG by moving to a multimodal representation.
For Figure~\ref{fig:pathological} (c) a text model for the query ``wedding band" ranked the left and right listing 4 positions apart, while the multimodal model ranked the right listing 427 positions higher. ``wedding band" saw a $\mathbf{1.55\%}$ \textbf{increase} in NDCG by moving to a multimodal representation.

%-------------------------------------------------------------------------
\section{Conclusion}
\label{sec:conclusion}
Learning to rank search results is one of Etsy and other e-commerce sites' most fundamental problems. In this paper we describe how deep visual semantic features can be transferred successfully to multimodal learning to rank framework. We verify in a large-scale experiment that there is indeed significant complementary information present in images that can be used to improve search ranking quality. We visualize concrete examples of this marginal value: (i) the ability of the image modality to capture style information not well described in text. (ii) The ability of the image modality to disentangle highly different listings considered similar under a text-only modality.

%-------------------------------------------------------------------------
\section*{Acknowledgments}
The authors would like to thank Arjun Raj Rajanna, Christopher Kanan, Robert Hall, and Will Gallego for fruitful discussions during the course of this paper.

{\small
\bibliographystyle{ieee}
\bibliography{etsy_arxiv_2015}
}

\end{document}